\documentclass{article}

\usepackage{url}
\usepackage[hidelinks]{hyperref}
\usepackage{graphicx}
\usepackage{amsmath}

\title{On educating machine learners}

\author	{George Leu and Jiangjun Tang}

\begin{document}

\maketitle

\begin{abstract}
Machine education is an emerging research field that focuses on the problem which is inverse to machine learning. To date, the literature on educating machines is still in its infancy. A fairly low number of methodology and method papers are scattered throughout various formal and informal publication avenues, mainly because the field is not yet well coalesced (with no well established discussion forums or investigation pathways), but also due to the breadth of its potential ramifications and research directions. In this study we bring together the existing literature and organise the discussion into a small number of research directions (out of many) which are to date sufficiently explored to form a minimal critical mass that can push the machine education concept further towards a standalone research field status.

\end{abstract}

\section{Introduction}

With the unprecedented achievements in machine learning concepts and technologies, machine agents are getting closer and closer to true autonomous operation in all respects implied by the definition of autonomy \cite{Abbass2016}. %In the last few decades a significant shift took place from classic symbolic rational reasoning towards learning and adaptation as the underpinnings of machine agents' intelligent behaviour. Several position papers in artificial intelligence \cite{Froese2009,Epstein2015} explain this shift in paradigm, noting the reliance of today's autonomous/intelligent machine agents on adaptation and learning, or more generally, on skill acquisition processes.
As learning machine agents gained momentum, a problem that is inverse to learning started to become increasingly relevant. This is the education problem, where if a learning mechanism is in place, the quality of the training that the learner undertakes is paramount for the learning outcome. In other words, if two agents implement the same learning mechanism, the one that receives \textit{better education} will perform better the tasks of interest. Thus, educating a machine learner is an attempt to find an optimal training program (i.e. the way the training inputs are organised), to maximise certain outcomes of the learning process in relation to the tasks of interest.

The attempt to organise these inputs is in many ways similar to designing a curriculum or an educational program in human education, which is perhaps the best analogy that can be used to convey the machine education idea to an audience. The analogy is nevertheless forced, because is implies somewhat that today's machine agents possess sufficient learning capabilities to be subject to education in similar ways the humans are, yet, it is useful because is very intuitive. In the case of humans, education is possible due to the fact that all ingredients/enablers of long-term continual learning are in place, i.e. the bio-physiological substrate consisting of memory and computation capabilities along with the relevant sensing and embodiment. In the case of machines, we can endeavour to speak about education in the context of long-term continual learners, where this can be seen as a facilitator for achieving artificial general intelligence (strong AI). However, we must be aware of the limitations, which are still major, machine learners have in comparison to human learners. Thus, while the concept of machine education is the most pertinent in relation to artificial general intelligence, the pioneering work reported in the literature in the last decade started from the narrow artificial intelligence concept (weak AI), where curriculum design is applied to single task (or task-set) machine learners used in general problem-solving.

Certainly, both strong and weak AI endeavours can benefit greatly from the machine education concept. However, the challenge when trying to discuss the machine education field is the great variety of learning machine agents, in terms of both the learning mechanisms and the embodiment they employ (that includes disembodied agents too). Inevitably, there is no one education framework that can be envisaged to fit all, and hence, we should expect educational endeavours to be in great extent empirical and context-related, at least in these early days of the machine education research. Thus, in this study, rather than discussing particular approaches, which are indeed very diverse in the issues they attempt to tackle, we consider beneficial to organise the discussion based on four major directions of research in machine education, which already attracted sufficient interest and generated a body of literature which, while still small, is sufficient to create a momentum that moves forward the whole field. These directions account for the educational goal, or in other words, the learning outcome that the educator wants to maximise, and are as follows: improving the learning speed~\cite{Bengio2009}, improving the task proficiency~\cite{Kumar2010}, achieve trustworthy AI~\cite{Leu2017a}, and achieve strong AI~\cite{Thorisson2015}.

We believe that this article has the merit of bringing together for the first time studies in machine education that are greatly scattered throughout many publication avenues, with many of them being relatively informal. %We also believe that the timing of this survey, the venue, and the AI audience it is intended for are paramount for the advancement of the field.

In light of the above, the discussion is structured as follows. Section~\ref{Section:History} presents a brief historical view that captures the emergence of the machine education concept. Then, Sections~\ref{Section:Faster},~\ref{Section:Better},~\ref{Section:Trust},~and~\ref{Section:AGI} bring together and discuss the existing work in each of the four directions. In the end, Section~\ref{Section:Conclusions} summarises and concludes the article.

%Thus, this paper aims to generate the foundation of the machine education concept, and establish the main directions of investigation that enable this concept to contribute towards building trusted interaction within heterogeneous teams (and, ultimately, a society) of machines and humans. We believe that machine education and the associated skill-based developmental view are relevant to multiple applications, such as, but definitely not limited to, surveillance, exploration or disaster recovery, where human-machine teaming is the pertinent mean of achieving present and future goals in a society.

\section{A Historical View} \label{Section:History}

% https://arxiv.org/pdf/1801.05927.pdf

The idea that education can be used in computational domain to facilitate artificial intelligent behaviour preceded even the actual computer technology. In his seminal study on computers and intelligence, published in 1950~\cite{Turing2009}, Turing, among other ideas, speculated on a way to achieve machine intelligence in a non-deterministic manner, without employing either logic or knowledge~\cite{Koza2003}. This is the education and learning way---that is, growing/nurturing the intelligence, rather than building it, in an machine agent.

While learning has been extensively addressed since then, the education side received attention only very recently, even though in essence, any machine learning piece of research contains implicitly, in a certain extent, a necessary insight into the inverse problem too.

%and mainly in empirical studies, with little efforts to formalise the process.
% https://www.aaai.org/ocs/index.php/AAAI/AAAI15/paper/viewFile/9487/9685
% http://www.jmlr.org/papers/volume17/15-630/15-630.pdf
The first study that explicitly treated the problem of educating machines was published by Bengio and colleagues~\cite{Bengio2014} in 2009 in a weak AI context, where the authors present a method to organise the sequence in which the data samples from the training set are presented to a neural learner, with the purpose of accelerating learning. The approach to education is coined by the authors as curriculum learning, a name that later has been used in several other studies in weak AI contexts. This was an important milestone for machine education, and, arguably though, can be considered the debut of research in this field.

%machine teaching problem formulation has been recently studied in the context of diverse applications including personalized educational systems, cyber-security problems, robotics, program synthesis, human-in-the-loop systems, and crowdsourcing. [Jha et al. 2016; Zhu 2015; Mei & Zhu 2015; Ba & Caruana 2014; Patil et al. 2014; Singla et al. 2014; Cakmak & Thomaz 2014] 

Another milestone was the first attempt to establish a comprehensive formalism for the education problem, which was published by Zhu in 2015~\cite{Zhu2015}. Several studies prior to that, including Bengio's curriculum learning study from 2009, formalised the education concept; however, the respective formalisms were context-related and applicable only to the specific example problems the authors wanted to investigate. In the 2015 study Zhu collates some of the previous attempts into a general mathematical formalism where the education process is described as a tandem between a teaching dimension and a learning dimension. From the learning agent's perspective, a training set $D$ is given, and the agent learns a model $\hat{\theta}$. The learning process can be then expressed using the regularised empirical risk minimisation framework given in Equation~\ref{Equation:Learning}:
\begin{equation}
	\hat{\theta} = argmin_{\theta \in \Theta} \sum_{(x,y) \in D} l(x,y,\theta) +\lambda||\theta||^2
	\label{Equation:Learning}		
\end{equation}
where $\Theta$ is the hypothesis space, $l$ is the loss function, $D$ is the training set, and $\lambda$ is the regularisation weight.
Education assumes the opposite process, in which the target model $\theta^*$ is given, and the educator has to find a training set $D$ which will lead the learner trained on $D$ to (approximately, according to \cite{Zhu2015}) learn $\Theta^*$. Then, education becomes a bi-level optimisation problem in which Equation~\ref{Equation:TeachingL1} is the education objective (educator's problem), and Equation~\ref{Equation:TeachingL2} is the learner's machine learning problem. The educator's goal is to bring learner's learned model $\hat{\Theta}$ close to the target $\Theta^*$ while keeping the teaching set at a feasible/minimal size ($||D||$ is the cardinality of the teaching set).
\begin{equation}
	\min_{D,\hat{\theta}} ||\hat{\theta} - \theta^*||^2 + \eta||D||_0
	\label{Equation:TeachingL1}
\end{equation}
subject to
\begin{equation}
	\hat{\theta} = argmin_{\theta \in \Theta} \sum_{(x,y) \in D} l(x,y,\theta) +\lambda||\theta||^2 
	\label{Equation:TeachingL2}		
\end{equation}

The formalism captured well the concept of machine education in a weak AI perspective, but made some limiting assumptions; for example, the educator (1) optimises over a discrete space of teaching sets and (2) needs to know the learning algorithm (model) to formulate the optimisation. The latter especially can be argued with a counter-example described in \cite{Leu2017}, which shows that an autonomous educator implemented via an evolutionary computation algorithm can find an optimal training set for a learner with no knowledge about how the learner operates. In order to address some of the limitations, Zhu et al. later extended the formalism to include more education goals~\cite{Zhu2018}; however, this is outside of the scope of this article.
%Since these formalisms have not yet gained great visibility and momentum in the literature, not many discussions exist about how they can be instantiated for practical applications, or whether they hold for the great diversity of machine learning and education problems. However, we believe they create a good picture of the machine education formal landscape, and can be used as a substrate for the next sections, where the most important machine education purposes, and consequently research directions, are discussed.

In 2017, the third and perhaps most important milestone has been reached when the first meeting dedicated to machine education took place at NIPS'17, under the name ``\textit{NIPS'17 Workshop: Teaching machines, robots and humans}''. This was the first event where machine education researchers brainstormed to identify challenges and opportunities of the emerging field and put the basis of a research community. The work presented and discussed at the meeting consisted of a number of position papers that speculated on potential research directions in machine education and sketched several investigation pathways to be followed to advance knowledge in the field. %A summary of the discussions can be found in a post-workshop article~\cite{Zhu2018}, where among other important aspects related to educating machines, the potential directions of research are also captured.

This study builds on the milestones mentioned above, and discusses four of the research directions discussed at the 2017 machine education meeting: improve the speed of learning, improve the task competency, achieve trustworthy AI, and achieve strong AI. While more potential investigation pathways could be brought into attention, some are still in a speculative stage with no concrete work reported in the literature so far, therefore they do not provide a basis for discussion as yet.

\section{Education for Faster Learning} \label{Section:Faster}

When the learning goal is defined, and the amount of learning is fixed, speeding up the learning process, or learning from less samples can be greatly beneficial. In contexts where large models need to be learn, computation cost, in terms of either or both time and processing, is one of the major bottlenecks. Optimisation of the learning cost has been addressed under the concept of \textit{curriculum learning}, starting with the pioneering work of Bengio et al. \cite{Bengio2009}. Like all efforts related to educating machines, curriculum learning optimises the environmental and peer inputs to the agent, such as the sequence, quantity and composition of the corresponding training data, except that the typical goal is narrowed down to reducing the cost of learning.

In general, curriculum learning is used in stationary\footnote{Recently, non-stationary contexts have been also approached, where reinforcement learning has been treated from a machine education perspective ~\cite{Florensa2017,Svetlik2017,Clayton2019}. However, research in this direction is still in a very incipient stage.} supervised learning contexts, and most of the work is related to neural (non-symbolic) learners, even though in theory it can be applied to any type of learning (for example, a curriculum design applied to a learner with both symbolic and non-symbolic components can be found in \cite{Leu2016}). An entirely non-symbolic approach to curriculum learning is presented in~\cite{Bengio2009}, where the authors use deep neural networks that learn initially from simple examples and then progress towards learning from more complex ones in order to gradually increase proficiency in a task of interest. In this way, a skill bootstrap effect is obtained, but this is not emphasised by the authors. The claimed purpose is to speed up the learning, based on the assumption that by choosing the order in which examples are presented to the learner, one can guide training towards faster learning. In~\cite{Bengio2009} the authors see a curriculum as a sequence of training iterations. For each training criterion in the sequence a certain set of weights is assigned to the training examples. This can be generalised as a process of re-weighting the training distribution. In the initial stage (first training iteration), the weights favour training examples that describe the simplest concepts and can be learnt easy. The next training iterations involve slight changes in the weighting of examples, which increase at each step the chances of sampling slightly more difficult examples. At the end of the sequence, the re-weighting of the examples leads to a uniform weight distribution, which means the training is performed on the target training set/distribution. Starting from this, Bengio and colleagues continued investigating the rationality underlying this learning paradigm in a number of studies, and discussed the relationship between conventional optimisation techniques and the educational approach, i.e. the continuation method and the annealing method. More details about these methods can be found in \cite{Bengio2013} and \cite{Bengio2014a}.

The curriculum learning concept has been instantiated by other authors too, in a range of applications where the key is to find a ranking function that assigns learning priorities to training samples. In a more recent study \cite{Zaremba2015}, Zaremba and Sutskever refine the approach presented in \cite{Bengio2009}, and use a similar curriculum learning method in conjunction with a Recurrent NN (RNN) with long short-term memory (LTSM) units. In \cite{Khan2011} the authors attempt to teach a robot the concept of \textit{graspability}, which involves deciding if an object can be grasped and picked up with one hand. In this study human subjects were asked to assign a learning sequence of graspability to various objects, and ranking was determined by the common sense of the participants.
 
In curriculum learning, the curriculum, which can be seen as a ranking function, is typically derived by predetermined heuristics for particular problems. An example is in the original curriculum learning study of Bengio et al. \cite{Bengio2009} where the authors apply the methodology to classifying geometrical shapes. The ranking function in this case is derived by the variability in shape, which means the shapes exhibiting less variability should be learned earlier. In \cite{Spitkovsky2009} the authors used grammar induction, where the ranking function is derived in terms of the length of a sentence. In this study the heuristic is that the number of possible solutions grows exponentially with the length of the sentence. Thus, shorter sentences are easier to learn and should be learnt earlier.

The heuristics involved in curriculum learning is beneficial, as it implies flexibility to incorporate prior knowledge from various sources~\cite{Jiang2015}. However, this may lead to inconsistency between the fixed curriculum and the dynamically learned models. This equates with a curriculum that is predetermined \textit{a priori} and cannot be adjusted accordingly to take into account any feedback about the learner. From a human behavioural perspective, evidence have shown that curriculum learning is consistent with some of the principles in human teaching \cite{Khan2011}, i.e. the teacher driven education.

\section{Education for Increased Competency} \label{Section:Better}

Improving the level of task competency the learner achieves at the end of the learning process is another purpose of the educational effort. This aspect of machine education has been investigated under several approaches grouped under the umbrella of \textit{self-paced learning} (SPL) and \textit{self-paced curriculum learning} (SPCL). Both follow the same approach to education like curriculum learning, where the instructional sequence allows the learner to gradually progress from easy to more complex samples in training. However, they target improved task proficiency, i.e. the learning outcome, rather than the optimisation of the instructional process (faster learning).

Self-paced learning, introduced by Kumar et al. \cite{Kumar2010}, approaches the organisation of training experience from easy to complex from the perspective of the learner, which in human education would be similar to student-driven education. This approach starts from the fact that a readily computable measure of the complexity of training experiences may not be available to the teacher. Hence, ordering the training sequence from easy to more difficult prior to the beginning of the training, like in curriculum learning, may be challenging if not impossible. The solution proposed by SPL is to decide the sample order based on the feedback received from learner, as training progresses through the training iterations. This means the curriculum is determined dynamically during the training to adjust to learner's learning pace. Thus, at each iteration the self-paced learning algorithm simultaneously selects easy samples and learns a new parameter vector. The number of samples selected is governed by a weight that is annealed until the entire training data has been considered.

An application of SPL can be found in \cite{Lee2011}, where Lee and Grauman use the concept for a visual discovery task. Their method, called \textit{self-paced discovery}, learns to categorise objects by extracting compact, object-level models from a pool of unlabelled image data. The method focuses on the easiest instances first, and progressively expands its repertoire to include more complex objects. Easier regions are defined as those with both high likelihood of generic \textit{objectness} and high familiarity of surrounding objects. At each cycle of the discovery process, the easiness of each sub-window in the pool of images is re-evaluated, and a single prominent cluster is then retrieved from the easiest instances. As the system gradually accumulates models, each new and more difficult discovery benefits from the knowledge provided by earlier discoveries. The study demonstrates the superiority of the self-paced method over the traditional machine learning approach, where batch learning over randomised training sequence is used. In batch learning, where no educational process is employed, equal attention is paid to all training instances, simultaneously, regardless of the strength of their appearance or context cues.

Curriculum learning and SPL generated substantial momentum in machine education. The two methods share a similar conceptual paradigm, but differ in the specific education schemes (teacher-student focus), as well as in the purpose (process-outcome improvement). In curriculum learning, the education scheme is predetermined by prior knowledge, and remains fixed throughout the learning process, being heavily dependant on the quality of the prior knowledge (or quality of the teacher) and ignoring feedback from the learner. In SPL, the education scheme is dynamically changed to fit the learning pace of the leaner, but the approach is unable to deal with prior knowledge. To overcome the issues posed by the two methodologies, a unified framework called \textit{self-paced curriculum learning} has been proposed in \cite{Jiang2015} by Jiang and colleagues.

SPCL is formulated as a concise optimisation problem that takes into account both prior knowledge known before training and the learning progress during training. In terms of similarities with human education, SPCL would account for a collaborative teacher-student educational approach. The education scheme learns latent variable models through an iterative approach where at each iteration two actions are performed: selection of easy samples and update of the parameter vector. The number of samples selected at each iteration is determined by a weight that is gradually annealed such that later iterations introduce more samples. The algorithm converges when all samples have been considered and the objective function cannot be improved further.

A form of self-paced curriculum learning is presented in \cite{Romero2014} by Romero et al. as a knowledge distillation problem, where a small and fast neural learner uses the knowledge accumulated by a large and slow learner which plays the teacher role. Distillation is typically used in reinforcement learning contexts \cite{Rusu2015} (where it is known as policy distillation), and is not an educational process \textit{per se}, because the student learner only uses the final model learnt by the teacher network, in order to bootstrap certain skills. The main process employed in distillation is imitation learning; therefore, this is rather a traditional machine learning problem. However, in \cite{Romero2014}, the authors enhance distillation approach to obtain a cooperative teacher-student approach. In this approach, the student learner uses not only the final model but also intermediate representations learned by the teacher, dynamically, as required by the current training status. This allows the student to generalise better and also run faster. The study demonstrates using the CIFAR-10 dataset that a deep student network with almost 10.4 times less parameters outperforms in terms of both speed and learning outcome a larger, state-of-the-art teacher network.

\section{Education for Trustworthy AI} \label{Section:Trust}

With increased capabilities of the learning machine agents also comes the problem of their trust capabilities. These condition, or at least influence their social acceptance and inclusion, as well as the overall performance of the team/society they operate in. On the one hand, since they are inherently adaptive and operate based on learning from environmental and peer inputs, it becomes problematic whether they will achieve their intended purposes, and more importantly, whether they will do so without harming or impeding their peers (and the society at large) to operate efficiently. It is important thus that their behaviour is trustworthy from the perspective of others (peers, designers, etc.). On the other hand, while operating in a designated team for a particular task, or in society for general purposes, these agents should also be able to trust/distrust their peers and take from them or give them control over tasks, whenever needed, to achieve (either individual or collective) goals. Trust is therefore at the core of social intelligent behaviour \cite{Abbass2016a}, and from a machine intelligence perspective this is a fundamental issue in relation to intelligent agents, which are supposed to operate in real-life conditions, in our society. Socially, heterogeneous teams of human and machine agents require first of all that trust between team members is enabled at each moment in time \cite{Abbass2016}. Without trust, delegation of tasks, acceptance of tasks, sharing of tasks and any collaborative intelligent behaviour in general may be hardly achievable, and thus, the functionality of the team may be further affected, with significant impact on its success.

From a trust perspective, through a curriculum, machine education focuses on enabling the machine agents to achieve training-induced value systems \cite{Merrick2010} that guide them to behave in the intended way and not otherwise. While not entirely excluding hard constraints and prescriptive behaviours, in machine education it is typically desired that agents gain trustworthy use of their skills purely by extracting the value systems through exposure to external world and learning. In human-machine teaming, trust is concerned not only with gaining an agent's own trustworthy behaviour, but also with the agent gaining trust in behaviours of others in the team \cite{Abbass2016a}. Machine education can contribute to this when the system of values of an agent emerges from observing both physical environment and peers agents. Therefore, an agent can be educated or can self-educate for both sides of trustworthiness: become trustworthy, and trust others (humans or machines); in addition, the agent can be re-educated or can self-educate continuously towards expansion of its skills and value system, in order to be reused in different missions and/or different teams. In this manner, the machine education paradigm can be used to create trustworthy machine learning systems.

The use of education with a focus on trust is still in its infancy, with very few studies grouped typically around the idea of designing curricula to overcome adversarial influence (i.e. adversarial sampling). Curriculum design for adversarial contexts has been studied mainly with a focus on adversarial alterations of the training data, known as \textit{causative availability} \cite{Wang2010}, where alterations can be done based on various sample selection methods. Examples are targeted alterations focusing on particular samples that are of interest for an adversary, or indiscriminate alterations produced with randomly selected samples across the training data set. Regardless of the way the samples to be altered are chosen, the alteration itself plays an important role in learning accuracy and subsequent trustworthiness of the learner agent \cite{Leu2017}.
Thus, the pertinent way to ensure these agents achieve the intended goals, is to train them using appropriate exposure to or interaction with relevant training material, or in other words, to find optimal curricula through which the agents can be educated to perform well even in the presence of adversarial activity. The educational process in this case is concerned with finding the appropriate sequence, quantity and composition of the training data, so that the learner can learn properly regardless of the adversarial inputs \cite{Wang2010,Papernot2016}.

Adversarial influence \cite{Moosavi-Dezfooli2016} can manifest in the training samples, in the test samples, or in both. Depending on this, robustness of a learning agent can be discussed from several points of view. The learning point of view is concerned with finding the learning mechanism itself, which can guarantee the desired behaviour of the learning agents \cite{Gu2014,Shaham2015}. The testing point of view is focused on ensuring the appropriate trade-off between generalisation and overfitting is achieved, so that the learner does not respond to stimuli outside a desired range \cite{Goodfellow2014a}.

Adversarial influence has perhaps the most serious effects on neural learners, due to well known issues of artificial neural networks, which suffer greatly from perturbations in the input data. The input perturbation issue has been first investigated by Szegedy et al. \cite{Szegedy2013}, who showed that even for the most advanced deep neural networks, which generalise very well on recognition/classification tasks, small imperceptible non-random perturbations can lead to significant changes in network's prediction and accuracy. Later work of \cite{Gu2014} extended the study and showed for an image processing task that such distortions can result in $100\%$ misclassification for a state of the art deep neural network. The method they proposed for mitigating these effects employed both on the education (pre-processing and training strategies) and learning mechanisms (network topology) through the addition of a denoising autoencoder. A more recent study investigated the nature of the distorsions, proposing a linear formalisation, and further studied their mathematical underpinning and propagation through the representations build by networks in their various layers \cite{Goodfellow2014a}. In a different study Papernot et al. \cite{Papernot2016} proposed a curriculum based on distillation as a defence mechanism against adversarial perturbations in deep neural networks, while Shaham and colleagues \cite{Shaham2015} attempted to increase the local stability of neural learners through a curriculum based on Robust Optimisation. Later, in~\cite{Leu2017}, evolutionary computation has been used to evolve a curriculum that takes advantage of the forgetting mechanisms inherent to deep neural learners. The adversarial influence in this study is of the categorical type, where a fraction of the training samples is intentionally labelled with erroneous categories. A supervised learning agent in the form of a deep convolutional neural network which learns to recognise/classify handwritten digits from $0$ to $9$, and an mutation-only genetic algorithm is used to evolve the order of the data samples in the training set (including the adversarial samples) to mitigate the influence of the adversarial samples on learning accuracy.

\section{Education for Strong AI} \label{Section:AGI}

Of all research directions discussed in this article, AGI is the least covered in the literature. While the AGI field itself has sufficient breadth and depth, as well as a long tradition, it focuses on the learning problem, even though education is very tightly coupled to it and inherently embedded. Thus, out of this literature we could select and discuss only a small number of studies which are explicitly positioned as contributions to the education problem.

Strong AI has been traditionally associated with the concepts of \textit{lifelong machine learning} in machine learning, and \textit{continual cognitive development} in cognitive science. In both directions the key characteristics of continual development are continuous learning process, knowledge and skill acquisition, and use of past experience to facilitate future learning \cite{Chen2016a}. Through these, the purpose is to exceed the isolated learning paradigm on which machine learning has been founded, and achieve complex cognitive machine agents that operate similar to humans. Designing education programs for strong AI has been only very recently addressed, and focused in two directions: top-down and bottom-up development.

In the top-down perspective the investigations focus on tasks, and typically mix cognitive task analysis~\cite{Leu2016a} and system dynamics~\cite{Gonzalez2007}; i.e. they design frameworks for facilitating the learning of known complex tasks by decomposing and organising then into simple primitive tasks. Several studies \cite{Thorisson2015,Thorisson2016} discussed the need of a task theory in relation to artificial systems, to handle the developmental perspective. Thorrisson et al. \cite{Thorisson2016} position the task concept at the core of artificial systems, due to the fact tasks are used for both training and evaluation of these systems, and therefore, tasks are fundamental for designing education programs. In \cite{Thorisson2015} a framework with 11 design principles is proposed in relation to both the tasks and the environment in which an artificial agent operates, with the purpose of facilitating a consistent path for development of strong AI capabilities. Later, in \cite{Thorisson2016} a framework with 6 design principles is proposed, taking into account only the tasks and not the environment. An application of such top-down approach in a practical context is reported in~\cite{Leu2016}, where the authors describe a curriculum applied to a task-based computational Sudoku solver. The solver learns to play Sudoku based on a set of primitive cognitive tasks which are learnt in isolation. The primitive tasks are then aggregated through a symbolic production system to solve Sudoku boards of a known difficulty in a cognitively plausible manner.

In the bottom-up perspective the focus is on skills, where the investigations involve machine agents or systems that acquire simple skills to perform simple tasks, and then gradually acquire through exposure to environment new and/or more complex skills. These allow the performance of new and/or more complex tasks which exceed the task-set envisaged by the educator at the design stage. Typically, the investigations employ non-symbolic approaches in the form of neural learning agents. %An example is to curriculum design is presented in \cite{Bengio2009} under the curriculum learning paradigm. The authors use deep neural networks that learn initially from simple examples and then progress towards learning from more complex ones in order to gradually increase proficiency in a task of interest. The purpose in this study however, is not to advance towards multiple task learning and skill transfer for generalisation, but rather to speed up the learning of a single task, based on the assumption that by choosing the order in which examples are presented to the learner, one can guide training towards better and faster learning. 
Such approach has been presented in a study by Zaremba and Sutskever~\cite{Zaremba2015}, who refine the classic curriculum learning approach introduced by Bengio~\cite{Bengio2009}. They use a similar curriculum learning method in conjunction with a recurrent neural network with long short-term memory units to provide the memory substrate needed for skill transfer and generalisation to new tasks. In another study \cite{Rusu2016}, the authors propose the \textit{progressive} neural networks, claiming advanced ability to gradually learn skills for solving multiple new tasks. They note that the proposed progressive networks are able to perform skill acquisition and transfer, while avoiding catastrophic forgetting. The proposed networks are tested in multiple reinforcement learning tasks on various 3D maze games, confirming that transfer occurs at both low-level sensory and high-level control layers of the learned policies.

\section{Conclusions} \label{Section:Conclusions}

This study presented a discussion on the literature related to the emerging field of machine education. The discussion brings together studies in machine education belonging to four major directions of research in this field. The four directions are: improve the learning speed, improve the task proficiency, achieve trustworthy AI, and achieve strong AI. These directions account for the educational goals, or in other words, the learning outcomes that the educator wants to maximise.

The discussions presented in this article show that machine education, while not yet coalesced as a standalone research field, has gained already sufficient momentum to be presented to an audience as a successful endeavour. Through this article, the author provides to the interested audience a valuable tool, i.e. a one-stop information source, to facilitate their own endeavours of pursuing one or more of the goals of machine education.

\bibliographystyle{unsrt}
\bibliography{TA_biblio,SkillBased,MemReviewBib}

% Generated by IEEEtran.bst, version: 1.14 (2015/08/26)
\begin{thebibliography}{10}
\providecommand{\url}[1]{#1}
\csname url@samestyle\endcsname
\providecommand{\newblock}{\relax}
\providecommand{\bibinfo}[2]{#2}
\providecommand{\BIBentrySTDinterwordspacing}{\spaceskip=0pt\relax}
\providecommand{\BIBentryALTinterwordstretchfactor}{4}
\providecommand{\BIBentryALTinterwordspacing}{\spaceskip=\fontdimen2\font plus
\BIBentryALTinterwordstretchfactor\fontdimen3\font minus
  \fontdimen4\font\relax}
\providecommand{\BIBforeignlanguage}[2]{{%
\expandafter\ifx\csname l@#1\endcsname\relax
\typeout{** WARNING: IEEEtran.bst: No hyphenation pattern has been}%
\typeout{** loaded for the language `#1'. Using the pattern for}%
\typeout{** the default language instead.}%
\else
\language=\csname l@#1\endcsname
\fi
#2}}
\providecommand{\BIBdecl}{\relax}
\BIBdecl

\bibitem{Abbass2016}
H.~Abbass, G.~Leu, and K.~Merrick, ``A review of theoretical and practical
  challenges of trusted autonomy in big data,'' \emph{IEEE Access}, vol.~4, pp.
  2808--2830, 2016.

\bibitem{Bengio2009}
Y.~Bengio, J.~Louradour, R.~Collobert, and J.~Weston, ``Curriculum learning,''
  in \emph{Proceedings of the 26th Annual International Conference on Machine
  Learning}, ser. ICML '09.\hskip 1em plus 0.5em minus 0.4em\relax New York,
  NY, USA: ACM, 2009, pp. 41--48.

\bibitem{Kumar2010}
\BIBentryALTinterwordspacing
M.~P. Kumar, B.~Packer, and D.~Koller, ``Self-paced learning for latent
  variable models,'' in \emph{Advances in Neural Information Processing Systems
  23}, J.~D. Lafferty, C.~K.~I. Williams, J.~Shawe-Taylor, R.~S. Zemel, and
  A.~Culotta, Eds.\hskip 1em plus 0.5em minus 0.4em\relax Curran Associates,
  Inc., 2010, pp. 1189--1197. [Online]. Available:
  \url{http://papers.nips.cc/paper/3923-self-paced-learning-for-latent-variable-models.pdf}
\BIBentrySTDinterwordspacing

\bibitem{Leu2017a}
G.~Leu, E.~Lakshika, J.~Tang, K.~Merrick, and M.~Barlow, ``Machine education -
  the way forward for achieving trust-enabled machine agents,'' in
  \emph{NIPS’17 Workshop: Teaching Machines, Robots, and Humans}, 2017.

\bibitem{Thorisson2015}
K.~R. Th{\'o}risson, J.~Bieger, S.~Schiffel, and D.~Garrett, \emph{Towards
  Flexible Task Environments for Comprehensive Evaluation of Artificial
  Intelligent Systems and Automatic Learners}.\hskip 1em plus 0.5em minus
  0.4em\relax Cham: Springer International Publishing, 2015, pp. 187--196.

\bibitem{Turing2009}
\BIBentryALTinterwordspacing
A.~M. Turing, \emph{Computing Machinery and Intelligence}.\hskip 1em plus 0.5em
  minus 0.4em\relax Dordrecht: Springer Netherlands, 2009, pp. 23--65.
  [Online]. Available: \url{https://doi.org/10.1007/978-1-4020-6710-5_3}
\BIBentrySTDinterwordspacing

\bibitem{Koza2003}
\BIBentryALTinterwordspacing
J.~R. Koza, M.~A. Keane, and M.~J. Streeter, ``Evolving inventions,''
  \emph{Scientific American}, vol. 288, no.~2, pp. 52--59, 2003. [Online].
  Available: \url{http://www.jstor.org/stable/26060164}
\BIBentrySTDinterwordspacing

\bibitem{Bengio2014}
Y.~Bengio, E.~Thibodeau-Laufer, G.~Alain, and J.~Yosinski, ``Deep generative
  stochastic networks trainable by backprop,'' in \emph{Proceedings of the 31st
  International Conference on Machine Learning}, Beijing, China, 2014.

\bibitem{Zhu2015}
X.~Zhu, ``Machine teaching: An inverse problem to machine learning and an
  approach toward optimal education,'' in \emph{Proceedings of the Twenty-Ninth
  AAAI Conference on Artificial Intelligence}, 2015.

\bibitem{Leu2017}
G.~Leu, J.~Tang, E.~Lakshika, K.~Merrick, and M.~Barlow, ``Curriculum
  optimisation via evolutionary computation, for a neural learner robust to
  categorical adversarial samples.'' in \emph{The 4th Asian Conference on
  Defence Technology}.\hskip 1em plus 0.5em minus 0.4em\relax IEEE Explore,
  2017.

\bibitem{Zhu2018}
\BIBentryALTinterwordspacing
X.~Zhu, A.~Singla, S.~Zilles, and A.~N. Rafferty, ``An overview of machine
  teaching,'' \emph{CoRR}, vol. 1801.05927, 2018. [Online]. Available:
  \url{http://arxiv.org/abs/1801.05927}
\BIBentrySTDinterwordspacing

\bibitem{Florensa2017}
\BIBentryALTinterwordspacing
C.~Florensa, D.~Held, M.~Wulfmeier, M.~Zhang, and P.~Abbeel, ``Reverse
  curriculum generation for reinforcement learning,'' in \emph{Proceedings of
  the 1st Annual Conference on Robot Learning}, ser. Proceedings of Machine
  Learning Research, S.~Levine, V.~Vanhoucke, and K.~Goldberg, Eds.,
  vol.~78.\hskip 1em plus 0.5em minus 0.4em\relax PMLR, 13--15 Nov 2017, pp.
  482--495. [Online]. Available:
  \url{http://proceedings.mlr.press/v78/florensa17a.html}
\BIBentrySTDinterwordspacing

\bibitem{Svetlik2017}
M.~Svetlik, M.~Leonetti, J.~Sinapov, R.~Shah, N.~Walker, and P.~Stone,
  ``Automatic curriculum graph generation for reinforcement learning agents,''
  in \emph{Proceedings of the 31st AAAI Conference on Artificial Intelligence},
  2017.

\bibitem{Clayton2019}
\BIBentryALTinterwordspacing
N.~Clayton and H.~Abbass, ``Machine teaching in hierarchical genetic
  reinforcement learning: Curriculum design of reward functions for swarm
  shepherding,'' \emph{CoRR}, vol. abs/1901.00949, 2019. [Online]. Available:
  \url{http://arxiv.org/abs/1901.00949}
\BIBentrySTDinterwordspacing

\bibitem{Leu2016}
G.~Leu and H.~Abbass, ``Computational red teaming in a sudoku solving context:
  Neural network based skill representation and acquisition,'' in
  \emph{Intelligent and Evolutionary Systems}.\hskip 1em plus 0.5em minus
  0.4em\relax Springer, 2016, pp. 319--332.

\bibitem{Bengio2013}
Y.~Bengio, A.~Courville, and P.~Vincent, ``Representation learning: A review
  and new perspectives,'' \emph{IEEE Transactions on Pattern Analysis and
  Machine Intelligence}, vol.~35, no.~8, pp. 1798--1828, Aug 2013.

\bibitem{Bengio2014a}
\BIBentryALTinterwordspacing
Y.~Bengio, \emph{Evolving Culture Versus Local Minima}.\hskip 1em plus 0.5em
  minus 0.4em\relax Berlin, Heidelberg: Springer Berlin Heidelberg, 2014, pp.
  109--138. [Online]. Available:
  \url{https://doi.org/10.1007/978-3-642-55337-0_3}
\BIBentrySTDinterwordspacing

\bibitem{Zaremba2015}
W.~Zaremba and I.~Sutskever, ``Learning to execute,'' \emph{arXiv:1410.4615v3},
  2015.

\bibitem{Khan2011}
S.~Khan, Y.~Zou, A.~Amjad, A.~Gardezi, C.~L. Smith, C.~Winters, and T.~S.
  Reese, ``Sequestration of camkii in dendritic spines in silico,''
  \emph{Journal of Computational Neuroscience}, vol.~31, no.~3, pp. 581--594,
  2011.

\bibitem{Spitkovsky2009}
V.~Spitkovsky, H.~Alshawi, and D.~Jurafsky, ``Baby steps: How less is more
  in unsupervised dependency parsing,'' in \emph{NIPS: Grammar Induction,
  Representation of Language and Language Learning}, 2009.

\bibitem{Jiang2015}
L.~Jiang, D.~Meng, Q.~Zhao, S.~Shan, and A.~G. Hauptmann, ``Self-paced
  curriculum learning.'' in \emph{Proceedings of the Twenty-Ninth AAAI
  Conference on Artificial Intelligence}, 2015.

\bibitem{Lee2011}
Y.~J. Lee and K.~Grauman, ``Learning the easy things first: Self-paced visual
  category discovery,'' in \emph{CVPR 2011}, June 2011, pp. 1721--1728.

\bibitem{Romero2014}
\BIBentryALTinterwordspacing
A.~Romero, N.~Ballas, S.~E. Kahou, A.~Chassang, C.~Gatta, and Y.~Bengio,
  ``Fitnets: Hints for thin deep nets,'' \emph{CoRR}, vol. abs/1412.6550, 2014.
  [Online]. Available: \url{http://arxiv.org/abs/1412.6550}
\BIBentrySTDinterwordspacing

\bibitem{Rusu2015}
\BIBentryALTinterwordspacing
A.~A. Rusu, S.~G. Colmenarejo, {\c{C}}.~G{\"{u}}l{\c{c}}ehre, G.~Desjardins,
  J.~Kirkpatrick, R.~Pascanu, V.~Mnih, K.~Kavukcuoglu, and R.~Hadsell, ``Policy
  distillation,'' \emph{CoRR}, vol. abs/1511.06295, 2015. [Online]. Available:
  \url{http://arxiv.org/abs/1511.06295}
\BIBentrySTDinterwordspacing

\bibitem{Abbass2016a}
H.~A. Abbass, E.~Petraki, K.~Merrick, J.~Harvey, and M.~Barlow, ``Trusted
  autonomy and cognitive cyber symbiosis: Open challenges,'' \emph{Cognitive
  computation}, vol.~8, no.~3, pp. 385--408, 2016.

\bibitem{Merrick2010}
K.~Merrick, ``A comparative study of value systems for self-motivated
  exploration and learning by robots,'' \emph{IEEE Transactions on Autonomous
  Mental Development}, vol.~2, no.~2, pp. 119--131, 2010.

\bibitem{Wang2010}
\BIBentryALTinterwordspacing
S.~L. Wang, K.~Shafi, C.~Lokan, and H.~A. Abbass, ``Adversarial learning: the
  impact of statistical sample selection techniques on neural ensembles,''
  \emph{Evolving Systems}, vol.~1, no.~3, pp. 181--197, Oct 2010. [Online].
  Available: \url{https://doi.org/10.1007/s12530-010-9013-y}
\BIBentrySTDinterwordspacing

\bibitem{Papernot2016}
N.~Papernot, P.~McDaniel, X.~Wu, S.~Jha, and A.~Swami, ``Distillation as a
  defense to adversarial perturbations against deep neural networks,'' in
  \emph{2016 IEEE Symposium on Security and Privacy (SP)}, May 2016, pp.
  582--597.

\bibitem{Moosavi-Dezfooli2016}
S.-M. Moosavi-Dezfooli, A.~Fawzi, and P.~Frossard, ``Deepfool: A simple and
  accurate method to fool deep neural networks,'' in \emph{The IEEE Conference
  on Computer Vision and Pattern Recognition (CVPR)}, June 2016.

\bibitem{Gu2014}
\BIBentryALTinterwordspacing
S.~Gu and L.~Rigazio, ``Towards deep neural network architectures robust to
  adversarial examples,'' \emph{CoRR}, vol. abs/1412.5068, 2014. [Online].
  Available: \url{http://arxiv.org/abs/1412.5068}
\BIBentrySTDinterwordspacing

\bibitem{Shaham2015}
\BIBentryALTinterwordspacing
U.~Shaham, Y.~Yamada, and S.~Negahban, ``Understanding adversarial training:
  Increasing local stability of neural nets through robust optimization,''
  \emph{CoRR}, vol. abs/1511.05432, 2015. [Online]. Available:
  \url{https://arxiv.org/abs/1511.05432}
\BIBentrySTDinterwordspacing

\bibitem{Goodfellow2014a}
\BIBentryALTinterwordspacing
I.~J. Goodfellow, J.~Shlens, and C.~Szegedy, ``Explaining and harnessing
  adversarial examples,'' \emph{CoRR}, vol. abs/1412.6572, 2014. [Online].
  Available: \url{http://arxiv.org/abs/1412.6572}
\BIBentrySTDinterwordspacing

\bibitem{Szegedy2013}
\BIBentryALTinterwordspacing
C.~Szegedy, W.~Zaremba, I.~Sutskever, J.~Bruna, D.~Erhan, I.~Goodfellow, and
  R.~Fergus, ``Intriguing properties of neural networks,'' \emph{CoRR}, vol.
  abs/1312.6199, 2013. [Online]. Available:
  \url{http://arxiv.org/abs/1312.6199}
\BIBentrySTDinterwordspacing

\bibitem{Chen2016a}
Z.~Chen and B.~Liu, ``Lifelong machine learning,'' \emph{Synthesis Lectures on
  Artificial Intelligence and Machine Learning}, vol.~10, no.~3, pp. 1--145,
  2016.

\bibitem{Leu2016a}
G.~Leu and H.~Abbass, ``A multi-disciplinary review of knowledge acquisition
  methods: From human to autonomous eliciting agents,'' \emph{Knowledge-Based
  Systems}, vol. 105, pp. 1 -- 22, 2016.

\bibitem{Gonzalez2007}
\BIBentryALTinterwordspacing
C.~Gonzalez and V.~Dutt, ``Learning to control a dynamic task: A system
  dynamics cognitive model of the slope effect,'' in \emph{Proceedings of the
  8th International Conference on Cognitive Modeling}.\hskip 1em plus 0.5em
  minus 0.4em\relax Taylor and Francis/Psychology Press, 2007, pp. 61--66.
  [Online]. Available: \url{http://repository.cmu.edu/sds/75}
\BIBentrySTDinterwordspacing

\bibitem{Thorisson2016}
K.~R. Th{\'o}risson, J.~Bieger, T.~Thorarensen, J.~S. SigurÃ°ard{\'o}ttir,
  and B.~R. Steunebrink, \emph{Why Artificial Intelligence Needs a Task
  Theory}.\hskip 1em plus 0.5em minus 0.4em\relax Cham: Springer International
  Publishing, 2016, pp. 118--128.

\bibitem{Rusu2016}
A.~Rusu, N.~Rabinowitz, G.~Desjardins, H.~Soyer, J.~Kirkpatrick,
  K.~Kavukcuoglu, R.~Pascanu, and R.~Hadsell, ``Progressive neural networks,''
  \emph{CoRR}, vol. abs/1606.04671, 2016.

\end{thebibliography}

\end{document}